# Federated Reinforcement Learning for Runtime Optimization of AI Applications in Smart Eyewears

Hamta Sedghani
*Dipartimento di Elettronica*
*Informazione e Bioingegneria*
*Politecnico di Milano*
Milan, Italy
hamta.sedghani@polimi.it

Abednego Wamuhindo Kambale
*Dipartimento di Elettronica*
*Informazione e Bioingegneria*
*Politecnico di Milano*
Milan, Italy
abednegowamuhindo.kambale@polimi.it

Federica Filippini
*Dipartimento di Elettronica*
*Informazione e Bioingegneria*
*Politecnico di Milano*
Milan, Italy
federica.filippini@polimi.it

Francesca Palermo
*EssilorLuxottica*
*Smart Eyewear Lab*
Milan, Italy
francesca.palermo@luxottica.com

Diana Trojaniello
*EssilorLuxottica*
*Smart Eyewear Lab*
Milan, Italy
diana.trojaniello@luxottica.com

Danilo Ardagna
*Dipartimento di Elettronica*
*Informazione e Bioingegneria*
*Politecnico di Milano*
Milan, Italy
danilo.ardagna@polimi.it

*Abstract*—Extended reality technologies are transforming fields such as healthcare, entertainment, and education, with Smart Eye-Wears (SEWs) and Artificial Intelligence (AI) playing a crucial role. However, SEWs face inherent limitations in computational power, memory, and battery life, while offloading computations to external servers is constrained by network conditions and server workload variability. To address these challenges, we propose a Federated Reinforcement Learning (FRL) framework, enabling multiple agents to train collaboratively while preserving data privacy. We implemented synchronous and asynchronous federation strategies, where models are aggregated either at fixed intervals or dynamically based on agent progress. Experimental results show that federated agents exhibit significantly lower performance variability, ensuring greater stability and reliability. These findings underscore the potential of FRL for applications requiring robust real-time AI processing, such as real-time object detection in SEWs.

## I. INTRODUCTION

Artificial Intelligence (AI) has demonstrated transformative potential across diverse fields such as healthcare, agriculture, and entertainment by enhancing decision-making and operational efficiency [1], [2]. Integrating AI into end-user devices, particularly Smart Eye-Wear (SEW), enables real-time environmental interaction and processing. These devices often execute computationally intensive computer vision tasks such as classification, object detection, and segmentation using deep neural networks (DNNs). However, the limited processing power, battery life, and memory of SEW devices pose significant challenges to executing complex AI models locally [3], [4]. To address these limitations, DNN partitioning has emerged as a practical solution, allowing computational workloads to be split across SEW devices, smartphones, and cloud servers [5]–[8]. This strategy facilitates adaptive offloading of heavy processing tasks, improving performance while managing energy consumption and latency.

Reinforcement Learning (RL) is a promising approach for managing DNN partitioning and offloading decisions due to its inherent ability to learn optimal policies through interaction with a dynamic environment [9], [10]. In SEW and other mobile AI systems, the runtime conditions, such as network bandwidth, cloud latency, battery level, and CPU load, can vary unpredictably. Static or heuristic-based partitioning strategies often fail to adapt effectively to such variability, leading to suboptimal performance. In contrast, RL agents continuously observe the system state, take actions (e.g., decide where to execute each layer of the DNN), and receive feedback in the form of rewards (e.g., lower latency, reduced energy consumption, fewer constraint violations). Over time, the agent learns to map system states to optimal actions, enabling it to make real-time, context-aware decisions [10], [11]. Building upon this, Federated Reinforcement Learning (FRL) [12] extends traditional reinforcement learning by enabling multiple agents to collaboratively learn optimal policies without sharing raw interaction data. This distributed approach offers several key advantages over standard RL, particularly in privacy-sensitive and resource-constrained environments. Each agent independently interacts with its environment, learns from it, and shares an update, which may include revised value function estimates, model weights, or other components such as gradients. FRL preserves user privacy by keeping data localized, reduces communication overhead by transmitting only model updates, and improves scalability across distributed networks. Furthermore, by aggregating knowledge from diverse local experiences, FRL enhances generalization and learning efficiency, even under non-independent and identically distributed data conditions commonly found in real-world applications [12].

In our previous work [11], we demonstrated the efficacy of RL-based DNN partitioning solutions that reduce energy use, 5G network fees, and latency by selecting configurations adaptively. This work extends the solution proposed in [11] and introduces a three-tier, layer-wise DNN partitioning framework paired with FRL-based runtime offloading strategy. It adapts to fluctuating 5G/WiFi bandwidth and cloud latency to extend SEW battery life while preserving responsive execution. The study explores the effects of FRL on optimizing SEW resource allocation, offering insights for more robust and scalable AI solutions in wearable devices. We performed an evaluation of the proposed solution by considering network bandwidth and cloud latency variability. Results indicate response time constraint violation below 5% on average, demonstrating particularly FRL is well-suited for resource-constrained devices such as SEWs and smartphones operating in dynamic, privacy-sensitive environments.

This paper is organized as follows. Related works are discussed in Section II. Section III describes the problem statement while Section IV presents our FRL-based approach. In Section V the results of the conducted experiments are discussed, while conclusions are finally drawn in Section VI.

## II. Related Work

Task offloading and DNN partitioning have been extensively explored to overcome the computational and energy constraints of resource-limited devices such as smartphones and SEWs [11], [13]. These methods distribute workloads across edge devices, smartphones, and cloud servers to optimize latency, energy consumption, and overall performance.

A seminal contribution in this area is Neurosurgeon [13], which proposes fine-grained, layer-level DNN partitioning between mobile devices and the cloud. Its lightweight scheduler selects optimal partition points based on static performance models to balance latency and energy. While effective, Neurosurgeon lacks adaptive decision-making and does not leverage learning-based strategies to respond to dynamic runtime conditions. Authors in [14] proposed a joint task partitioning and offloading strategy for DNN-enabled Mobile Edge Computing (MEC) networks. Their approach incorporates a layer-level partitioning mechanism that enables mobile devices to execute DNN tasks locally or offload them to edge servers. To optimize scheduling and resource allocation, the framework integrates a delay prediction model and a dynamic pricing mechanism. Leveraging game-theoretic distributed algorithms, the method reduces processing delay and energy consumption, outperforming traditional methods in MEC environments. Similarly, [15] proposed a Stackelberg game-based strategy for DNN-based task partitioning and offloading in MEC environments tailored to real-time AI applications. Their approach employs a three-tier model partitioning strategy, distributing computations across local devices, edge servers, and cloud virtual machines. By modeling the MEC platform as a leader and mobile users as followers, the framework enables adaptive cloud-side resource provisioning that responds to dynamic user demands and computational workloads. Likewise, authors in [16] present a multi-objective optimization framework for task partitioning and offloading in real-time computer vision applications. They propose a RL-based strategy integrated with Lyapunov optimization to jointly balance computational efficiency and inference accuracy.

Recently, FRL has gained traction as an effective approach for offloading DNN-based tasks in decentralized environments. For instance, [17] proposed a FRL framework for dynamic task offloading in hierarchical IoT networks. Their approach employs DRL to optimize resource allocation while minimizing energy consumption and latency. A global agent aggregates gradients from local devices via weighted averaging to update a central policy, which is redistributed for further local training. Similarly, the authors of [18] introduced a FRL framework for task offloading in digital twin-enabled edge networks. Their two-layer architecture includes a physical layer for task execution and a digital twin layer for modeling real-time system states. Local DRL agents make offloading decisions, while a global agent at the base station aggregates their gradients to coordinate resource allocation and configure passive reflecting surfaces. This design improves efficiency and reduces overall system cost. In a related direction, [19] addresses federated learning for edge caching by combining DRL with hierarchical model aggregation. Cluster heads perform weighted averaging of local models and disseminate updated parameters back to devices. This method supports scalable collaboration and accommodates heterogeneous data distributions and device capabilities.

In this work, we propose a three-tier, layer-level DNN partitioning framework combined with a FRL-based runtime offloading strategy. Our approach accounts for dynamic variations in 5G and WiFi throughput, as well as cloud latency, aiming to extend SEW battery life while maintaining acceptable end-to-end execution time to ensure a seamless user experience. To the best of our knowledge, this is the first study to address this problem using FRL in the context of DNN partitioning and task offloading for smart wearable devices.

## III. Problem Statement

This section introduces the reference scenario underpinning our study, as detailed in Section III-A, followed by the problem mathematical formulation presented in Section III-B.

### A. Reference Scenario

We consider the same system model as in our previous work [11], where a SEW device runs AI applications powered by DNNs for real-time tasks such as object detection, tracking, and classification. The system consists of three main entities (see Figure 1): the SEW device, a paired smartphone, and a cloud server. To ensure a smooth user experience, we impose execution time constraints—e.g., an end-to-end latency below 33 ms to maintain the frame rate above 30 fps [20].

The SEW, equipped with a CPU, neural processing unit, and limited battery, is connected via WiFi to a smartphone that may offload computation to a cloud server over a 5G connection. Although 5G mmWave offers high bandwidth, it suffers from significant variability [21]. Similarly, WiFi throughput can fluctuate due to interference from nearby devices [22], and cloud servers may experience queuing delays due to dynamic workloads [23]. While our system adapts to such variability through local decision-making, cloud resource allocation remains beyond our control. Depending on the current system conditions (battery level, network throughput, or latency), the computation can be partitioned at layer granularity and distributed dynamically across the three tiers. This is managed by a RL agent deployed on the phone, which selects from a set of feasible DNN configurations the one that minimizes energy consumption and 5G usage cost while meeting latency constraints. The agent makes decisions periodically, adapting to dynamic conditions such as throughput fluctuations (WiFi interference, 5G variability) and cloud server delays.

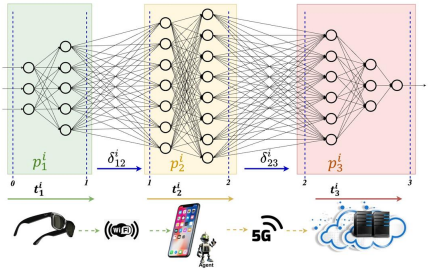

Figure 1: Reference system including a SEW, mobile phone and cloud.

## B. Mathematical Formulation

In this work, we consider image-based AI applications that process requests supplied in the form of image frames with a frequency $\lambda$ (expressed in frames per second). All the parameters we will introduce in the following are measured in each time window of duration $\tau$. Let $\mathcal{K}$ denote the set of candidate DNN configurations, where each configuration $\kappa^i \in \mathcal{K}$ consists of three partitions. A special partition $p_\emptyset^i$ indicates no execution on a given device. Partition-to-device mapping is fixed: $p_1^i$ runs on the SEW, $p_2^i$ on the smartphone, and $p_3^i$ on the cloud. If a device is not involved in execution, its partition is set to $p_\emptyset^i$. Data transmission between partitions is represented by $\delta_{dk}^i$ (in bytes), with $\delta_{12}^i$ and $\delta_{23}^i$ denoting data transfers from SEW to smartphone and smartphone to cloud, respectively. When computation is fully local, $\delta_{12}^i = \delta_{23}^i = 0$, and when fully offloaded, both equal the input tensor size $\delta_0$. Given the device assignments, execution latencies $t_1^i, t_2^i, t_3^i$ for each partition are determined via DNN profiling as in [13].

To manage the execution of AI tasks on SEW devices, we define a decision-making framework that selects an optimal DNN configuration, either running the entire model locally or offloading specific segments to the smartphone or cloud server. This selection is modeled as an optimization problem aimed at minimizing the combined energy cost on the SEW and the monetary cost of 5G data transmission:

$$\min_x \left( \alpha (E_{SEW} + E_{phone}) + c_{5G} \right) \tau \quad \text{(P1a)}$$

subject to:

$$l_{total} < L_{max}, \quad \text{(P1b)}$$

$$\sum_{i=1}^{|\mathcal{K}|} x^i = 1. \quad \text{(P1c)}$$

$E_{SEW}$ denotes the local energy consumption to run $p_1^i$ on the SEW ($e_{local\_SEW} = \tau \lambda \sum_{i=1}^{|\mathcal{K}|} z_{SEW} x^i$) and the energy required to transmit intermediate data $\delta_{12}$ to the smartphone over WiFi ($e_{tr\_SEW} = \tau \lambda \sum_{i=1}^{|\mathcal{K}|} \theta_{SEW} \delta_{12}^i x^i / r_{WIFI}$). $E_{phone}$ indicates the local energy consumption to run $p_2^i$ on the phone ($e_{local\_phone} = \tau \lambda \sum_{i=1}^{|\mathcal{K}|} z_{phone} x^i$) and the energy required to transmit intermediate data $\delta_{23}$ to the smartphone over WiFi ($e_{tr\_phone} = \tau \lambda \sum_{i=1}^{|\mathcal{K}|} \theta_{phone} \delta_{23}^i x^i / r_{WIFI}$). The variables $z_{SEW}$ and $z_{phone}$ quantify the electrical energy needed by the SEW and phone, respectively, to perform one FLOP, while $\theta_{SEW}$ and $\theta_{phone}$ indicate the electrical power required by their network interface while sending data. The binary variable $x^i$ flags the selected DNN configuration. The cost $c_{5G}$ is given by $c_{5G} = \lambda g \sum_{i=1}^{|\mathcal{K}|} \delta_{23}^i x^i$, where $g$ is the cost for sending one byte of data from the mobile phone to the cloud using the 5G connection. $l_{total}$ is the total execution time, given by $l_{total} = l_{SEW} + l_{phone} + l_{sp} + l_{pc} + l_{cloud}$ where $l_{SEW} = \sum_{i=1}^{|\mathcal{K}|} t_1^i x^i$ and $l_{phone} = \sum_{i=1}^{|\mathcal{K}|} t_2^i x^i$ represent the execution time of partition $p_1^i$ on the SEW and $p_2^i$ on the mobile phone, respectively, $l_{sp} = \sum_{i=1}^{|\mathcal{K}|} \delta_{12}^i x^i / r_{WIFI}$ and $l_{pc} = \sum_{i=1}^{|\mathcal{K}|} \delta_{23}^i x^i / r_{5G}$ are the time to transfer $\delta_{12}$ from SEW to the phone and $\delta_{23}$ from the phone to the cloud. $l_{cloud}$ is the cloud execution time, which depends on $t_3^i$ and the current workload on the cloud server (which is unknown). The latency threshold $L_{max}$ in constraint (P1b) ensures real-time responsiveness required for an acceptable user experience. The optimization goal is to minimize operational costs over a control interval $\tau$, where $\alpha$ is the cost per joule. Note that, while cloud execution delays are included, cloud service costs are excluded from the optimization, aligning this work with a user-centric perspective. Although the optimization at a given time slot $\tau$ can be solved through exhaustive search over all configurations with a complexity of $O(|\mathcal{K}|)$, this approach fails to account for temporal variations in system performance. In contrast, RL enables the agent to model and anticipate these dynamics, allowing for more adaptive and energy-efficient decisions.

We formulate the offloading problem as a Markov Decision Process (MDP) defined by the tuple $\langle \mathbf{S}, \mathbf{A}, P, c, \gamma \rangle$, where $\mathbf{S}$ is the (possibly infinite) set of states; $\mathbf{A}(s)$ is the finite set of actions available in state $s$; $P(s'|s, a)$ is the transition probability from state $s$ to $s'$ after action $a$; $c(s, a, s')$ is the immediate cost of taking action $a$ in state $s$ leading to state $s'$; $\gamma \in [0, 1]$ is a discount factor that balances costs over time.

We define the agent state as $s = (r_{WIFI}, r_{5G}, l_{SEW}, l_{phone}, l_{cloud})$ where $r_{WIFI}, r_{5G}$ and $l_{cloud}$ are exogenous variables observed from the environment and not influenced by the agent actions. These represent network conditions and cloud latency, which vary independently. In contrast, the latencies of the SEW and smartphone ($l_{SEW}, l_{phone}$) depend on the selected action but are assumed to be known in advance, as execution times for all partitioning configurations can be estimated. Additionally, the smartphone is assumed to be under light load with no competing tasks.

The action space corresponds to the set of available DNN configurations, denoted by $|\mathcal{K}|$. We also include a special action $\eta$ to represent the case where the agent opts to keep the current configuration unchanged. Thus, $\mathbf{A}(s) = \{a^1, a^2, \ldots, a^{|\mathcal{K}|}\} \cup \{\eta\}$.

Due to stochasticity in network conditions and cloud latency, the system dynamics are unknown and cannot be modeled explicitly. Instead, we use observed transitions to guide learning. The cost function $c(s, a, s')$ aggregates multiple objectives: energy consumption in SEW and phone (as defined in [24]) $c_{SEW} = \alpha E_{SEW}\tau$, $c_{phone} = \alpha E_{phone}\tau$; 5G communication cost $c_{5G}(s, a)$; a penalty $c_{lat}(s, a, s') = \mathbb{1}_{\{l_{total} > L_{\max}\}}$ for violating latency constraints; a reconfiguration penalty $c_{cfg} = \mathbb{1}_{\{a \neq \eta\}}$ incurred when changing configuration. We combine the different costs using a simple additive weighting approach [25], defining $c(s, a, s')$ as:

$$c(s, a, s') = \omega_{SEW} \frac{c_{SEW}}{C_{SEW,max}} + \omega_{phone} \frac{c_{phone}}{C_{phone,max}} \quad (1)$$
$$+ \omega_{5G} \frac{c_{5G}}{C_{5G,max}} + \omega_{lat} c_{lat} + \omega_{rcfg} c_{rcfg}$$

Here, $\omega_{SEW}$, $\omega_{phone}$, $\omega_{5G}$, $\omega_{lat}$, and $\omega_{rcfg}$ are non-negative weights summing to one, and $c_{e_{SEW},max}$, $c_{e_{phone},max}$, and $c_{5G,max}$ are normalization factors for the SEW and phone energy costs and the 5G communication cost, respectively. The weights can be dynamically adjusted based on the SEW battery level—for instance, favoring local processing when energy is abundant, and prioritizing offloading as the battery depletes.

## IV. FEDERATED LEARNING SOLUTION

To tackle the problem described in Section III, we developed a federated variant of our reinforcement learning approach proposed in [11]. The federation mechanism is the core coordination process in FRL, enabling multiple reinforcement learning agents to collaboratively improve their policies by periodically sharing and aggregating model updates. Each agent is trained using the Deep Q-Network (DQN) algorithm, and two primary execution strategies are implemented: synchronous and asynchronous federation. A general view of our FRL solution is presented in Algorithm 1, which is implemented in RLlib [26] (version 2.12.0).

**Synchronous Execution**

In the synchronous mode, all agents complete the same number of training steps and contribute to the aggregation simultaneously. At each federation iteration, the master process initiates a training subprocess for each agent. These agents train independently for a fixed number of steps using their local environment and then return their updated model weights. The master process aggregates these weights by averaging them, producing a unified policy that is redistributed to all agents for the next iteration. This approach ensures consistency across agents but assumes they have similar training speeds and resources.

**Algorithm 1** Federated DQN Training Across $M$ Agents

**Require:** Federation iterations $N$, agents $M$, batch size $b$
1: Initialize global Q-network parameters $\theta^1$
2: **for all** agents $m = 1$ to $M$ **in parallel do**
3:     Initialize replay buffer $\mathcal{D}_m \leftarrow \emptyset$
4: **end for**
5: **for** $n = 1$ to $N$ **do**
6:     Master sends $\theta^r$ to all agents
7:     **for all** agents $m = 1$ to $M$ **in parallel do**
8:         Initialize Q-network $\theta_m \leftarrow \theta^n$
9:         **for** steps per training phase **do**
10:             **if** step=0 **then**:
11:                 Initialize agent
12:             **else**
13:                 Initialize agent's weights from the last aggregation this agent took part in
14:             **end if**
15:             Choose $a$ using $\epsilon$-greedy from $Q(s, a; \theta_m)$
16:             Take action $a$, observe $r$ and next state $s'$
17:             Store $(s, a, r, s')$ in $\mathcal{D}_m$
18:             Sample mini-batch of $b$ transitions from $\mathcal{D}_m$
19:             Compute target: $y = r + \gamma \max_{a'} Q(s', a'; \theta_m^-)$
20:             Perform gradient descent on $(y - Q(s, a; \theta_m))^2$
21:             $s \leftarrow s'$
22:         **end for**
23:         Agent sends updated $\theta_m$ to Master
24:     **end for**
25:     Master aggregates the weights using (2) or (3) to obtain $\theta^{n+1}$
26: **end for**
27: **return** Final global model $\theta^N$

**Asynchronous Execution**

To reflect real-world variability in resource availability and network conditions, we implement an asynchronous training strategy that removes the strict synchronization requirement of traditional federated learning. Since RLlib does not natively support such fine-grained asynchronous control, we designed and implemented a custom solution that emulates asynchronous execution. This approach gave us full control over agent behavior, enabling us to model realistic settings by configuring the number and behavior of fast and slow agents. Our framework thus offers a more flexible and scalable training process that better matches practical deployment scenarios. Agents are divided into fast and slow categories based on their training speeds. *Fast agents* complete training quickly and aggregate their updates together. In contrast, *slow agents* contribute their updates later and are incorporated individually into the existing aggregated model. Each iteration can thus involve multiple aggregations, one for fast agents and separate updates for each slow agent. Agents may switch between fast and slow roles dynamically based on observed conditions (e.g., network bandwidth). This design allows partial progress to be integrated without waiting for all agents, significantly improving flexibility and scalability.

**Aggregation Function**

The federation loop is coordinated by a central master process that, at each iteration, launches a subprocess for every participating agent and manages their execution flow. Each agent interacts with its local environment to collect data, performs independent training, and returns its updated policy weights to the master. These contributions are then combined using a dedicated aggregation function, which supports both synchronous and asynchronous execution models. In particular, the function allows agents with delayed training—the slow agents—to contribute as soon as their training concludes, ensuring their updates are seamlessly integrated.

The aggregation function processes the weight vectors produced by participating agents, incorporating, when applicable, the previously aggregated value and the corresponding number of contributing agents. It computes a normalized, weighted average over all inputs to produce a consistent global model.

Let the fast agents be indexed by $f \in \{1, \ldots, F\}$. Their contributions are aggregated as:

$$\theta_{\text{agg}}^{\text{fast}} = \frac{1}{F} \sum_{f=1}^{F} \theta_f. \quad (2)$$

Subsequently, as each slow agent $s \in \{1, \ldots, S\}$ completes training, its update is incorporated incrementally:

$$\begin{aligned} \theta_{agg}^0 &= \theta_{agg}^{fast} \\ \theta_{agg}^s &= \frac{(F+s-1)\theta_{agg}^{s-1} + \theta_s}{F+s}. \end{aligned} \quad (3)$$

Here, $\theta_i$ denotes the policy weights of agent $i$, and $\theta_{agg}^s$ is the aggregated model after incorporating slow agents up to index $s$. This recursive formulation ensures that late contributions are fairly integrated into the global model without disrupting the consistency of prior updates. From the second iteration onward, each agent reinitializes its policy using the most recent aggregated weights.

## V. EXPERIMENTAL RESULTS

This section presents a comprehensive evaluation of the proposed FRL approach across diverse scenarios. The experimental setup is described in Section V-A. Section V-B analyzes the impact of scaling the number of agents and compares our method against a state-of-the-art baseline. The performance of synchronous and asynchronous execution modes is assessed in Section V-C and Section V-D, respectively.

### A. Experimental Setup

We evaluate our framework using a low-frame-rate object detection application based on the widely adopted YOLOv5 model [27], [28], which offers 105 partitioning configurations. The SEW device is a Microsoft HoloLens 2 equipped with a Qualcomm Snapdragon 850 SoC, 4 GB RAM, and Adreno 630 GPU. The mobile tier comprises a Samsung S23 smartphone with a Snapdragon 8 Gen 2 SoC and 8 GB RAM, while the cloud layer is represented by a Dell Precision 5480 workstation featuring an Intel Core i7-13800H processor, 64 GB RAM, and an Nvidia RTX A1000 GPU. Devices communicate over WiFi 5 using the smartphone as a hotspot. To train the agent across all application instances, we utilized synthetic 5G and WiFi traces derived from real-world measurements, referred to as base traces. Given that the HoloLens supports WiFi5, all experiments in this section focus on WiFi5 scenarios. The WiFi trace, collected during profiling, consists of 3,000 samples at a 250 ms granularity. In contrast, the 5G trace includes 11,024 samples sourced from real-time measurements reported in [21]. To prevent overfitting to periodic patterns, each trace was replayed with randomized perturbations, including ±10% noise, temporal shifts, and mid-trace inversion. Additionally, for each cloud-executed partition $p_3^i$, cloud latency was sampled from an exponential distribution with rate parameter $\lambda = 1/t_3^i$, following [29]. Further details on the 5G, WiFi, and cloud latency traces are provided in the Appendix A.

The mobile phone runs an Android application implemented in Kotlin and C++, and the cloud server hosts a Python-based Flask container. We conduct a detailed profiling campaign to measure execution latency across partition points on all three devices, and energy consumption on the HoloLens and smartphone. Further details on partitioning parameters and profiling results are available in the Appendix B. We later extend our experiments to YOLOv8 [30], which introduces a deeper network and expands the configuration space to 210 partitioning options. The latency thresholds are set to 400 ms for YOLOv5 and 600 ms for YOLOv8. A reinforcement learning agent deployed on the smartphone makes task allocation decisions at 10-second intervals or, after five consecutive latency violations, every second until no violations are registered. In this experiment, our primary objective is to prevent execution time violations, which represent the most critical failure mode in the target system. To reflect this priority, we configure the cost function weights in Equation (1) as follows: $\omega_{lat} = 0.93$, $\omega_{e_{SEW}} = 0.03$, $\omega_{e_{phone}} = 0.02$, $\omega_{5G} = 0$, and $\omega_{rcfg} = 0.02$. This configuration places overwhelming emphasis on minimizing execution latency, particularly under the assumption that the SEW is fully charged, ensuring that energy and reconfiguration costs remain secondary to strict latency adherence. A summary of experimental parameters is provided in Table II in Appendix C.

An iteration in this context refers to the process leading up to the weight updates performed during federated learning. Specifically, each federation iteration encompasses agent initialization, training of each agent, aggregation, and weight updates. The total number of steps per agent across all iterations is set to 21,000 and the number of steps per agent, per iteration (updates frequency) is set to 500.

**Validation Methodology:** The validation plots presented in Sections V-C and V-D depict agent performance as a function of accumulated training experience, with the x-axis representing the number of training steps per agent. Each agent undergoes a validation phase comprising 300 steps after every 250 training steps. During each validation interval, violations are recorded and averaged. These average values are then plotted at intervals of 250 training steps, providing a clear view of performance progression while training progresses. For each experimental scenario, we conducted five random independent runs and reported both the mean and the range (minimum and maximum) of the observed values across these runs, offering a clear view of the results variability and robustness.

**Performance Metric**: We prioritized the most significant term in the cost function (1), namely, the violation of the end-to-end latency constraint (P1b). $c_{lat}(t) = \mathbb{1}_{\{l_{tot}(t) > L_{max}\}}$ (see Section III-B). Given the inherent noise in this metric, the training performance is assessed using the *moving average of violations*, with a sliding window of size $W = 1000$:

$$\text{MA}_{c_{\text{lat}}}(t) = \begin{cases} \frac{1}{t} \sum_{i=1}^{t} c_{\text{lat}}(i), & t < W, \\ \frac{1}{W} \sum_{i=t-W+1}^{t} c_{\text{lat}}(i), & t \geq W. \end{cases} \quad (4)$$

At the validation stage, violations are averaged over the validation period:

$$C_{\text{lat}}(k) = \frac{1}{V} \sum_{i=1}^{V} c_{\text{lat}}(kV + i), \quad (5)$$

where $V$ is the evaluation duration (300 steps), and $k$ is the index of the evaluation phase.

### B. Agent Scalability and comparison with Neurosurgeon

Our initial set of experiments investigated the impact of varying the total number of federated agents under synchronous execution. To this end, we trained a varying number of synchronous agents (10, 20, and 30) to optimize the runtime management of the *YOLOv5* AI application integrated into the SEW device and compared their performance against the baseline scenario represented by the single-agent case. As shown in Figure 2a, a satisfactory performance with a violation rate around 5%, are achieved after 9,000 steps, requiring approximately 2.5 hours of training. Although federated agents do not demonstrate faster learning compared to the single-agent case, they exhibit greater stability. As a baseline, we use Neurosurgeon [13], a leading partitioning algorithm that determines the optimal offloading point based on the previously observed network throughput. Since Neurosurgeon is originally limited to a single partition point—splitting the DNN between the mobile device and either an edge or cloud server—we adapt it to support double partitioning. This modification allows for the division of computation across three tiers, enabling a fair and consistent comparison with our proposed approach. While Neurosurgeon is designed to minimize either end-to-end latency or device energy consumption, it does not optimize for both metrics simultaneously. In Figure 2a, we observe that while Neurosurgeon (latency-optimized) effectively minimizes latency violations by predominantly selecting the local execution on SEW, it does so at the cost of significantly increased energy consumption in SEW—as confirmed by its elevated energy cost in Figure 2b. Conversely, Neurosurgeon (energy-optimized) achieves lower energy usage by favoring offloading to the phone and cloud but fails to respect latency constraints, resulting in a consistently high violation rate. This dichotomy highlights a core limitation of Neurosurgeon: its inability to jointly optimize latency and energy, as it targets only a single metric at a time. In contrast, our FRL approach achieves a better balance: it consistently maintains low latency violations and reduced energy consumption, especially with 20 and 30 agents. This effectiveness stems from our multi-objective reward function, which enables agents to adaptively optimize both goals simultaneously.

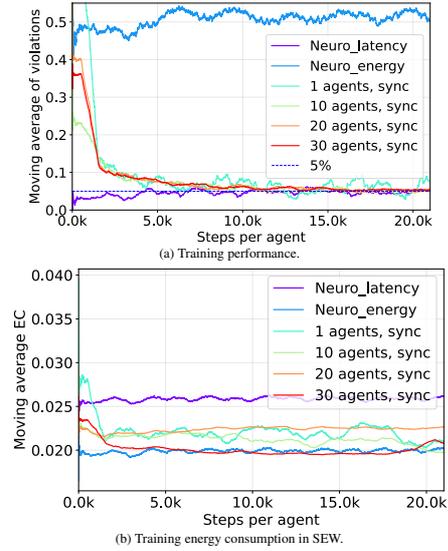

Figure 2: Synchronous agents trained on *YOLOv5*, $L_{max} = 400$, comparison with Neurosurgeon, a representative instance run.

### C. Policy Generalization and Application upgrade

This section explores the influence of the AI application deployed on the SEW device on the effectiveness and generalizability of FRL under synchronous execution. The goal is to assess whether policies trained within a FRL framework can generalize to slightly different applications. These differences arise not only from distinct time and energy profiles associated with each application, but also from mismatches in the number of partitioning points within the DNNs, resulting in varying numbers of configurations. To this end, we replace YOLOv5 object detection model with YOLOv8, a more complex DNN that introduces a broader configuration space and a more challenging decision-making environment for agents. A structured set of experiments is conducted to evaluate the adaptability and generalization capabilities of the learned policies:

1) **Training on YOLOv5 extended**: Agents are trained from scratch on an augmented version of YOLOv5 (called *YOLOv5 Extended*), in which the original configuration space is artificially expanded by duplicating existing configurations. This extension ensures a configuration space compatible with that of *YOLOv8*, with the maximum latency threshold set to $L_{max} = 600$.
2) **Transition to YOLOv8**: Agents are trained on the *YOLOv8* model, initialized from a policy pre-trained on *YOLOv5-Extended*, to evaluate transferability and adaptation in a more complex application setting.

The pretraining phase on YOLOv5 rapidly converges to minimal violation rates within approximately 2,000 train-ing steps (30 minutes) due to the relaxed latency threshold ($L_{max}$ = 600). Next, we evaluate the performance of federated training on the YOLOv8 application, where each agent policy is initialized from a model pre-trained on the extended YOLOv5 configuration space. This approach ensures consistency in action space dimensionality, facilitating a fair assessment of policy generalization. Figure 3 demonstrates that the agents achieve a 5% training violation rate within 1,000 steps (15 minutes), a substantial improvement compared to around 2.5 hours required when training from scratch (see Section V-B).

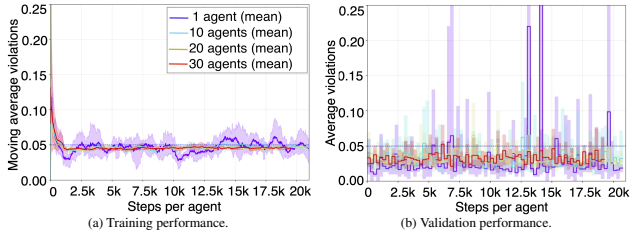

Figure 3: Synchronous agents trained on *YOLOv8*, $L_{max}$ = 600, 5 runs.

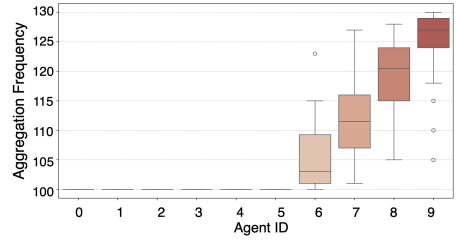

Figure 4: Distribution of Training Steps per Iteration, per Agent

### D. Asynchronous Execution

This section investigates the performance of agents trained using FRL under an asynchronous execution model, which reflects a general and realistic deployment scenario. In this setting, agents progress at different speeds, leading to unaligned aggregation times and introducing challenges in maintaining stable and effective learning. The AI application and latency constraint remain consistent with prior experiments, using *YOLOv5* on the SEW device with $L_{max}$ = 400 ms.

The asynchronous implementation introduces two key parameters that control asynchronicity:

- *proportion_slow_agents*: the fraction of agents that operate at a slower pace.
- *max_delay_slow_relative*: defines the maximum relative increase in training steps for slow agents, compared to the update frequency. While fast agents train for a fixed number of steps per iteration (*freq_updates*), slow agents perform a random number of steps between *freq_updates* and *freq_updates* * (1 + *max_delay_slow_relative*).

In each iteration, fast agents train and aggregate synchronously, receiving a shared update. Slow agents, however, complete training independently and receive personalized updates that do not influence previously aggregated models.

Figure 4 illustrates the training step distribution per agent in a scenario with 10 agents (6 fast, 4 slow), *proportion_slow_agents* = 40%, and *max_delay_slow_relative* = 30%. Agents are sorted based on their aggregation order, highlighting delays in update synchronization. To assess the impact of asynchrony, we systematically vary each parameter. First, we fix *max_delay_slow_relative* and vary the proportion of slow agents (Section V-D1). Then, we reverse the roles, holding the proportion constant while varying the delay parameter (Section V-D2). Each setting is evaluated for federations of 10 agents and compared against both single-agent baselines and synchronous federated training. Based on prior findings in Section V-B, the number of agents is capped at 20.

*1) Impact of the proportion of slow agents on performance*

This subsection evaluates the influence of the proportion of slow agents on the performance of asynchronous federated training. To isolate this factor, all other parameters are held constant while the proportion of slow agents is varied across 20%, 30%, and 40%. The analysis is conducted for federated setups comprising 10 agents, and for each configuration, we assess performance with *max_delay_slow_relative*=30%.

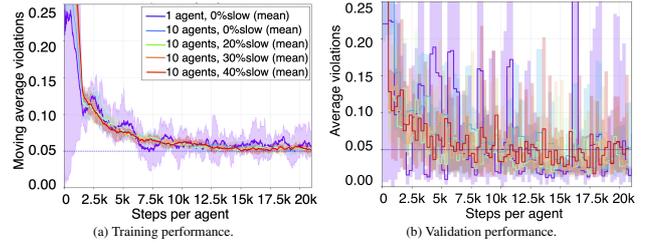

Figure 5: 10 Asynchronous agents with *max_delay_slow_relative*=30%

Results in Figure 5 indicates that both training and validation performance remain largely unaffected by changes in the proportion of slow agents. This suggests that federated learning in asynchronous settings maintains comparable effectiveness to the synchronous case, demonstrating robustness with respect to agent speed heterogeneity.

*2) Impact of training step increment for slow agents*

In this subsection, we examine how increasing the number of training steps assigned to slow agents affects the overall performance of asynchronous federated learning. Specifically, we evaluate the impact of maximum relative step increments of 10%, 20%, and 30%, involving 10 agents with *proportion_slow_agents*=40%. Findings indicate that increasing the training step increment for slow agents has minimal influence on both training and validation performance (see Figure 6). This suggests that the asynchronous federation mechanism is robust to moderate imbalances in per-agent training durations.

## VI. CONCLUSIONS

This study investigated a FRL framework for runtime DNN offloading in SEW devices, with a focus on adaptability, stability, and generalization. While FRL did not significantly accelerate convergence compared to single-agent training, it consistently produced more stable and robust policies. As shown in our experiments, federated agents exhibited lower variability during training and validation, particularly as the number of participating agents increased. This stability enables earlier training termination and contributes to more reliable deployment in real-time applications such as object detection and segmentation. The approach also demonstrated strong generalization across different AI applications (e.g., from YOLOv5 to YOLOv8) and remained effective under asynchronous execution, confirming its suitability for real-world, dynamic environments.

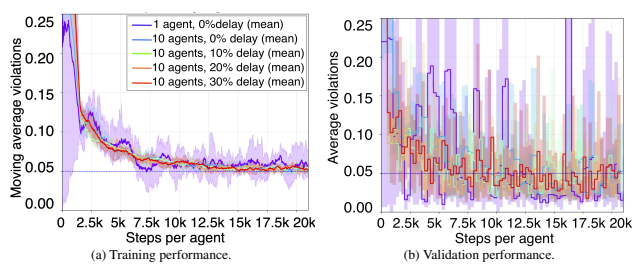

Figure 6: 10 Asynchronous agents, *proportion_slow_agents*=40%

Future work will explore advanced aggregation strategies, including weighted and hierarchical aggregation, to enhance learning efficiency and model personalization. Additionally, investigating gradient-based aggregation and extending the framework to policy-based RL methods such as PPO may offer improvements in convergence and scalability. These directions aim to further strengthen FRL utility in resource-constrained, privacy-sensitive edge AI applications.

## Acknowledgment

This paper was carried out in the EssilorLuxottica Smart Eyewear Lab, a Joint Research Center between EssilorLuxottica and Politecnico di Milano.

## References

[1] Y. He *et al.*, "Foundation model for advancing healthcare: Challenges, opportunities and future directions," *IEEE Reviews in Biomedical Engineering*, vol. 18, pp. 172–191, 2025.

[2] H. Sedghani *et al.*, "Advancing design and runtime management of ai applications with ai-sprint (position paper)," in *2021 IEEE 45th COMPSAC*, 2021, pp. 1455–1462.

[3] J. Coelho *et al.*, "Enabling Processing Power Scalability with Internet of Things (IoT) Clusters," *Electronics 2022, Vol. 11, Page 81*, vol. 11, p. 81, 1 2021.

[4] O. Debauche *et al.*, "A New Edge Computing Architecture for IoT and Multimedia Data Management," *Information*, vol. 13, p. 89, 2 2022.

[5] J. Karjee *et al.*, "Split computing: DNN inference partition with load balancing in IoT-edge platform for beyond 5G," *Measurement: Sensors*, vol. 23, p. 100 409, 2022.

[6] A. Parthasarathy *et al.*, "Partitioning and Placement of Deep Neural Networks on Distributed Edge Devices to Maximize Inference Throughput," in *32nd ITNAC*, IEEE, 2022, pp. 239–246.

[7] H. Sedghani *et al.*, "Space4ai-d: A design-time tool for ai applications resource selection in computing continua," *IEEE Transactions on Services Computing*, vol. 17, no. 6, pp. 4324–4339, 2024.

[8] N. Y. Yen *et al.*, "Partitioning DNNs for Optimizing Distributed Inference Performance on Cooperative Edge Devices: A Genetic Algorithm Approach," *Applied Sciences 2022, Vol. 12, Page 10619*, vol. 12, p. 10 619, 20 2022.

[9] S. Yuan *et al.*, "Joint optimization of dnn partition and continuous task scheduling for digital twin-aided mec network with deep reinforcement learning," *IEEE Access*, vol. 11, pp. 27 099–27 110, 2023.

[10] Z. Liu *et al.*, "Dnn partitioning, task offloading, and resource allocation in dynamic vehicular networks: A lyapunov-guided diffusion-based reinforcement learning approach," *IEEE Transactions on Mobile Computing*, vol. 24, no. 3, pp. 1945–1962, 2025.

[11] A. W. Kambale *et al.*, "Runtime management of artificial intelligence applications for smart eyewears," in *Proceedings of DML-ICC 2023*, Taormina, Italy: Association for Computing Machinery (ACM), 2023, pp. 1–8.

[12] A. Tuor *et al.*, "Federated reinforcement learning," *arXiv preprint arXiv:2102.08335*, 2021.

[13] Y. Kang *et al.*, "Neurosurgeon: Collaborative intelligence between the cloud and mobile edge," *SIGARCH Comput. Archit. News*, vol. 45, no. 1, pp. 615–629, 2017.

[14] M. Gao *et al.*, "Task partitioning and offloading in dnn-task enabled mobile edge computing networks," *IEEE Transactions on Mobile Computing*, vol. 22, no. 4, pp. 2435–2445, 2023.

[15] R. Sala *et al.*, "AI Applications Resource Allocation in Computing Continuum: A Stackelberg Game Approach," *IEEE Transactions on Cloud Computing*, vol. 13, no. 1, pp. 166–183, 2025.

[16] V. Sohn *et al.*, "Joint frame drop and object detection task offloading for mobile devices via rl with lyapunov optimization," *IEEE Transactions on Mobile Computing*, 2025.

[17] D. Kim *et al.*, "Collaborative policy learning for dynamic scheduling tasks in cloud–edge–terminal iot networks using federated reinforcement learning," *IEEE Internet of Things Journal*, vol. 11, no. 6, pp. 10 133–10 149, 2024.

[18] Y. Dai *et al.*, "Federated deep reinforcement learning for task offloading in digital twin edge networks," *IEEE Transactions on Network Science and Engineering*, vol. 11, no. 3, pp. 2849–2863, 2024.

[19] F. Majidi *et al.*, "Hfdrl: An intelligent dynamic cooperate cashing method based on hierarchical federated deep reinforcement learning in edge-enabled iot," *IEEE Internet of Things Journal*, vol. 9, no. 2, pp. 1402–1413, 2022.

[20] W. Luo *et al.*, "Multiple object tracking: A literature review," *Artificial Intelligence*, vol. 293, p. 103 448, 2021.

[21] A. Narayanan *et al.*, "A variegated look at 5g in the wild: Performance, power, and qoe implications," in *ACM SIGCOMM*, New York, NY, USA, 2021, pp. 610–625.

[22] T. Pulkkinen *et al.*, "Understanding WiFi Cross-Technology Interference Detection in the Real World," in *ICDCS*, 2020, pp. 954–964.

[23] D. Ardagna *et al.*, "Quality-of-service in cloud computing: modeling techniques and their applications," *J. Internet Serv. Appl.*, vol. 5, no. 1, 11:1–11:17, 2014.

[24] Z. Towfic *et al.*, "Benchmarking and testing of qualcomm snapdragon system-on-chip for jpl space applications and missions," in *IEEE Aerospace Conference (AERO)*, 2022, pp. 1–12.

[25] K. P. Yoon *et al.*, *Multiple attribute decision making: an introduction*. Sage publications, 1995.

[26] E. Liang *et al.*, "Rllib: Abstractions for distributed reinforcement learning," in *35th ICML*, 2018.

[27] J. Redmon *et al.*, "You only look once: Unified, real-time object detection," in *Proceedings of the IEEE Conference on Computer Vision and Pattern Recognition (CVPR)*, 2016, pp. 779–788.

[28] G. Jocher, *YOLOv5 by Ultralytics*, version 7.0, May 29, 2020. [Online]. Available: https://github.com/ultralytics/yolov5.

[29] U. Tadakamalla *et al.*, "Autonomic Resource Management for Fog Computing," *IEEE Transactions on Cloud Computing*, vol. 10, no. 4, pp. 2334–2350, 2022.

[30] G. Jocher *et al.*, *Yolo by ultralytics*, https://github.com/ultralytics/ultralytics, 2023.

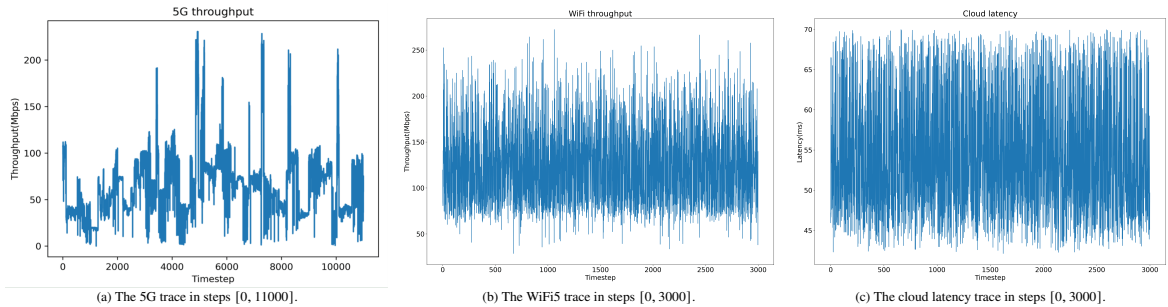

Figure 7: The snapshot of traces used for the experiments as the sources of variability in the system.

## APPENDIX A

To ensure that our DRL agent experiences realistic network variability during training, we incorporated empirical latency and throughput traces for 5G, WiFi, and cloud processing.

**5G Trace**: A real-world 5G uplink throughput trace was used to emulate cellular network variability [21]. The trace contains 11,024 measurements, with throughput ranging from 0 to 230.75 Mbps. To align with longer DRL training episodes (each with $4 \times 10^5$ steps), we periodically replayed the trace while applying randomized transformations — including circular shifts, ±10% Gaussian noise, and mid-point inversion — to avoid artificial periodicity. A representative snapshot is shown in Figure 7a.

**WiFi Trace**: WiFi performance was modeled using throughput traces collected during DNN profiling when the HoloLens SEW was connected to the smartphone via a local hotspot. This captures potential bandwidth fluctuations due to environmental interference or concurrent device usage. Similar to the 5G trace, the WiFi trace (3,000 steps) was extended via replay and randomized perturbations. Figure 7b illustrates a portion of the collected data.

**Cloud Latency Trace**: Cloud response time was modeled using an M/M/1 queuing system, assuming Poisson-distributed request arrivals — a standard approach in edge-cloud computing studies [29]. Execution latencies were sampled from exponential distributions whose means were configuration-dependent, based on profiling data (Table I). A trace representing the configuration in which the full DNN runs on the cloud is shown in Figure 7c.

Together, these traces introduce realistic, time-varying network and compute conditions that are essential for training and evaluating adaptive inference strategies in distributed environments.

## APPENDIX B

**YOLOv5 Partitioning and Profiling:** We employed the YOLOv5n object detection model in ONNX format to support cross-platform inference and eliminate framework dependencies. ONNX Runtime (v1.13.1) was used to execute various model partitions across three computational layers: (i) Microsoft HoloLens 2 as the Smart Edge Wearable (SEW), featuring a Snapdragon 850 SoC; (ii) Samsung Galaxy S23 (mobile device) with a Snapdragon 8 Gen 2; and (iii) a Dell Precision 5480 PC (cloud node) equipped with an Intel i7-13800HX CPU and an NVIDIA RTX A1000 GPU. These devices were connected via WiFi5 using the smartphone as a hotspot.

The ONNX format constrained partitioning to occur only at join nodes, resulting in 12 feasible partitioning points. To fully characterize edge-to-cloud configurations, we included two fictitious partitioning points to represent the entire model running on a single device (e.g., all on SEW, phone, or cloud), resulting in a total of 14 points. Using these, we systematically generated 105 unique configurations, covering all valid one- and two-stage partitioning combinations across the three layers. These include:

- 3 full-execution scenarios (model runs entirely on SEW, phone, or cloud),
- 36 one-split configurations (SEW–Phone, SEW–Cloud, Phone–Cloud), and
- 66 two-split configurations (SEW–Phone–Cloud).

Each configuration was profiled in terms of execution time, intermediate tensor sizes, and energy consumption. Execution times were averaged over five runs, with end-to-end latencies ranging from 220 ms to 580 ms depending on the configuration. Data transfer sizes for intermediate outputs between SEW–Phone and Phone–Cloud were also recorded, feeding into the runtime model.

To evaluate energy usage, we conducted a controlled profiling campaign using a TC66 power meter. We discharged the HoloLens and the phone from 100% to 95% while running a specific model partition, then measured the energy required to recharge from 95% to 100%. Repeating this for all 14 partition points allowed us to estimate the energy cost per iteration for each partition on both SEW and phone. To avoid exhaustive measurement in future evaluations, we trained a linear regression model to predict energy consumption based on two features: the number of floating-point operations (FLOPs) and the output tensor size. The model achieved a leave-one-out cross-validation error of only 5%, demonstrating high accuracy and generalizability.

In addition to compute energy, we profiled the network interface power consumption on both devices. To isolate this, we transmitted intermediate tensors of varying sizes without executing any DNN partition and measured energy consumption during transmission. This yielded average networking power values of 7.9 W on the HoloLens and 4.5 W on the S23, which were then used in the energy model for estimating the total cost of communication in hybrid deployments.

| Configuration parameters | | | | | | | |
|---|---|---|---|---|---|---|---|
| $\delta$(MB) | $\pi$(MB) | Partition 1 | | Partition 2 | | Partition 3 | |
| | | latency(ms) | $\mu$(MFLOPs) | latency(ms) | $\mu$(MFLOPs) | latency(ms) | $\mu$(MFLOPs) |
| (0, 6.25) | (0, 6.25) | (0, 82) | (0, 1680) | (0, 196) | (0, 5427) | (0, 78) | (0,5427) |

Table I: The range of DNN configuration parameters

The resulting profiling dataset — including latency, energy, FLOPs, and data transfer volumes — underpins our DRL agent's ability to make informed runtime decisions, optimizing the energy-latency trade-off across heterogeneous and variable deployment conditions.

## APPENDIX C

Table II outlines the key parameters used in our experiments.

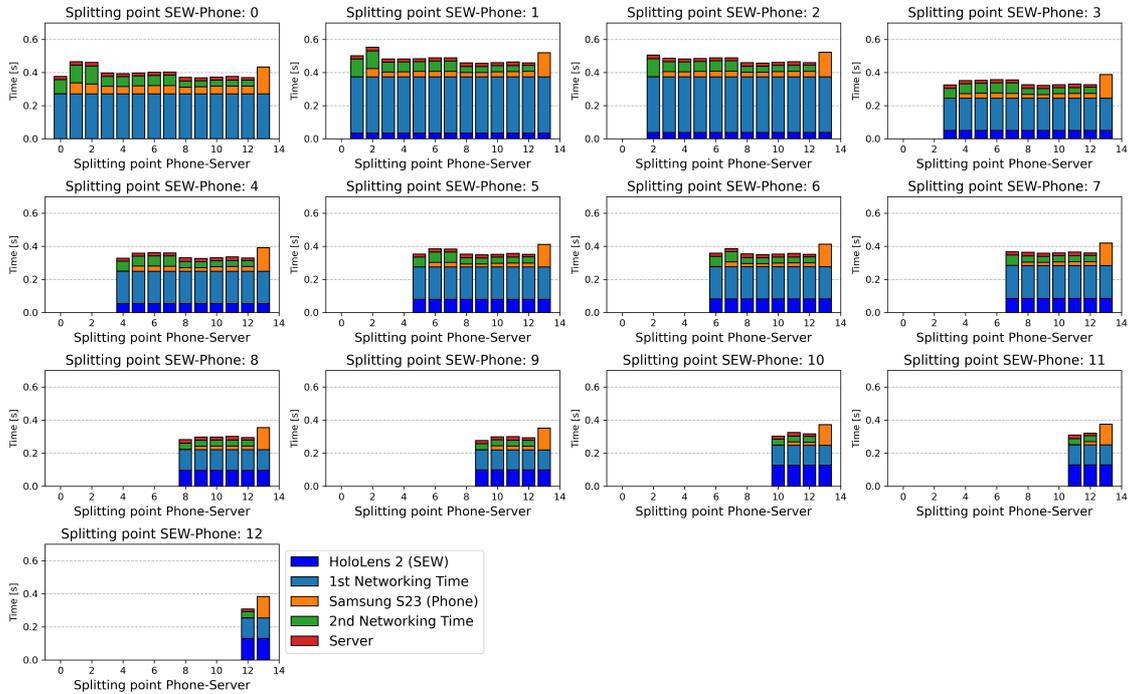

Figure 8: Profiling of the YOLOv5n on three devices: the SEW (HoloLens 2), the phone, and the cloud server.

| Reference scenario Parameters | |
|---|---|
| $|\mathcal{K}|$ (YOLOv5, YOLOv8) | (105, 210) |
| $g$ | 0.1 $/byte |
| Min and Max WiFi throughput value | [0, 580] byte/s |
| Min and Max 5G throughput value | [0, 350] byte/s |
| Min and Max SEW latency (YOLOv5, YOLOv8) | ([0, 450], [0, 660]) ms |
| Min and Max phone latency (YOLOv5, YOLOv8) | ([0, 65], [0, 110]) ms |
| Min and Max cloud latency (YOLOv5, YOLOv8) | ([0, 30], [0, 50]) ms |
| Min and Max percentage of battery | [10, 100]% |
| **Objective parameters** | |
| $\omega_{SEW}$ | 0.03 |
| $\omega_{phone}$ | 0.02 |
| $\omega_{5G}$ | 0.00 |
| $\omega_{lat}$ | 0.93 |
| $\omega_{rcfg}$ | 0.02 |
| $L_{max}$ (YOLOv5, YOLOv8) | (400, 600) ms |
| **Model hyper-parameters** | |
| Discout factor $\gamma$ | 0.99 |
| $\epsilon$ | 0.05 |
| $lr$ | 0.04 |
| Replay buffer | 10000 |
| Batch size | 512 |
| Number of layers | 3 |
| Number of neurons per layer | (100, 100, 60) |
| Activation function | ReLU |
| Dropouts | (0.4, 0.3, 0) |
| Discount factor $\gamma$ | 0.99 |
| Evaluation interval | 50 |
| Rollout fragment length | 5 |
| Target update frequency | 400 |
| Evaluation interval | 50 |
| **Federation parameters** | |
| Steps per agent | 21000 |
| Updates frequency | 500 |
| Number of agents | 1, 10, 20, 30 |

Table II: Experiments parameters.